\documentclass[10pt,twocolumn,letterpaper]{article}

\usepackage[pagenumbers]{cvpr}

\usepackage{xspace}
\usepackage{colortbl}
\usepackage{subcaption}

\usepackage{fontawesome}
\usepackage{academicons}

\usepackage{multirow}

\definecolor{red}{rgb}{0.8,0,0}
\definecolor{green}{RGB}{0, 133, 21}
\definecolor{grey}{rgb}{0.5,0.5,0.5}

\definecolor{w_1}{RGB}{52,204,204}
\definecolor{w_2}{RGB}{70,203,187}
\definecolor{w_3}{RGB}{95,202,161}
\definecolor{w_4}{RGB}{119,200,136}
\definecolor{w_5}{RGB}{155,199,101}
\definecolor{w_6}{RGB}{185,197,70}
\definecolor{w_7}{RGB}{227,194,28}
\definecolor{w_8}{RGB}{243,193,12}
\definecolor{w_9}{RGB}{255,192,0}

\newcommand{\ours}{\textbbf{\textsl{\textcolor{w_1}{W}\textcolor{w_2}{o}\textcolor{w_3}{r}\textcolor{w_4}{l}\textcolor{w_5}{d}\textcolor{w_6}{L}\textcolor{w_7}{e}\textcolor{w_8}{n}\textcolor{w_9}{s}}\nobreak\hspace{0.07em}}}

\newcommand{\textbbf}[1]{\scalebox{1.05}{\textbf{#1}}}

\definecolor{benchmarkgray}{rgb}{0.96, 0.96, 0.96}

\makeatletter
\def\blfootnote{\xdef\@thefnmark{}\@footnotetext}
\DeclareRobustCommand\onedot{\futurelet\@let@token\@onedot}
\def\@onedot{\ifx\@let@token.\else.\null\fi\xspace}
\def\eg{\textit{e.g}\onedot}

\makeatother

\usepackage{soul}

\definecolor{w_blue}{RGB}{52,204,204}
\definecolor{w_yellow}{RGB}{255,192,0}
\definecolor{cvprblue}{rgb}{0.21,0.49,0.74}

\usepackage[pagebackref=false,breaklinks,colorlinks,allcolors=cvprblue]{hyperref}

\usepackage[accsupp]{axessibility}

\def\paperID{}
\def\confName{CVPR}
\def\confYear{2026}

\title{Is Your Driving World Model an All-Around Player?\vspace{-0.1cm}}

\author{Lingdong Kong$^{*,1,\dagger}$,
Ao Liang$^{*,1}$,
Tianyi Yan$^{*,2}$,
Hongsi Liu$^{*,3}$,
Yu Yang$^{*,4}$,
Ziqi Huang$^{5}$,
Xian Sun$^{6}$,\\
Wei Yin$^{7}$,
Jialong Zuo$^{8}$,
Yixuan Hu$^{9}$,
Dekai Zhu$^{9}$,
Dongyue Lu$^{1}$,
Youquan Liu$^{10}$,
Guangfeng Jiang$^{3}$,\\
Linfeng Li$^1$,
Xiangtai Li$^5$,
Long Zhuo$^{11}$,
Lai Xing Ng$^{12}$,
Benoit R. Cottereau$^{13}$,\\
Changxin Gao$^{8}$,
Liang Pan$^{11}$,
Wei Tsang Ooi$^{1,\ddagger}$,
Ziwei Liu$^{5,\ddagger}$
\\[0.5ex]
{\small$^1$NUS}~~
{\small$^2$UM}~~
{\small$^3$USTC}~~
{\small$^4$ZJU}~~
{\small$^5$NTU}~~
{\small$^6$Duke}~~
{\small$^7$Horizon}~~
{\small$^8$HUST}~~
{\small$^9$TUM}~~
{\small$^{10}$FDU}~~
{\small$^{11}$SH Lab}~~
{\small$^{12}$A*STAR}~~
{\small$^{13}$CNRS}\\
{\small$^*$Equal Contributions}\quad
{\small$^\dagger$Project Lead}\quad
{\small$^\ddagger$Corresponding Authors}
\\[0.9ex]
\faGlobe~\textbf{Project Page:} \href{https://worldbench.github.io/worldlens}{\textcolor{w_blue}{\textbf{\textsl{Link}}}}~~\quad~~
\faGithubAlt~\textbf{GitHub:} \href{https://github.com/worldbench/WorldLens}{\textcolor{w_yellow}{\textbf{\textsl{Link}}}}~~\quad~~
\faGlobe~\textbf{HuggingFace:} \href{https://huggingface.co/datasets/worldbench/videogen}{\textcolor{w_blue}{\textbf{\textsl{Link}}}}
}

\begin{document}

\twocolumn[{
    \vspace{-0.1cm}
    \renewcommand\twocolumn[1][]{#1}
    \maketitle
    \begin{center}
    \centering
    \captionsetup{type=figure}
    \vspace{-0.7cm}
    \includegraphics[width=\textwidth]{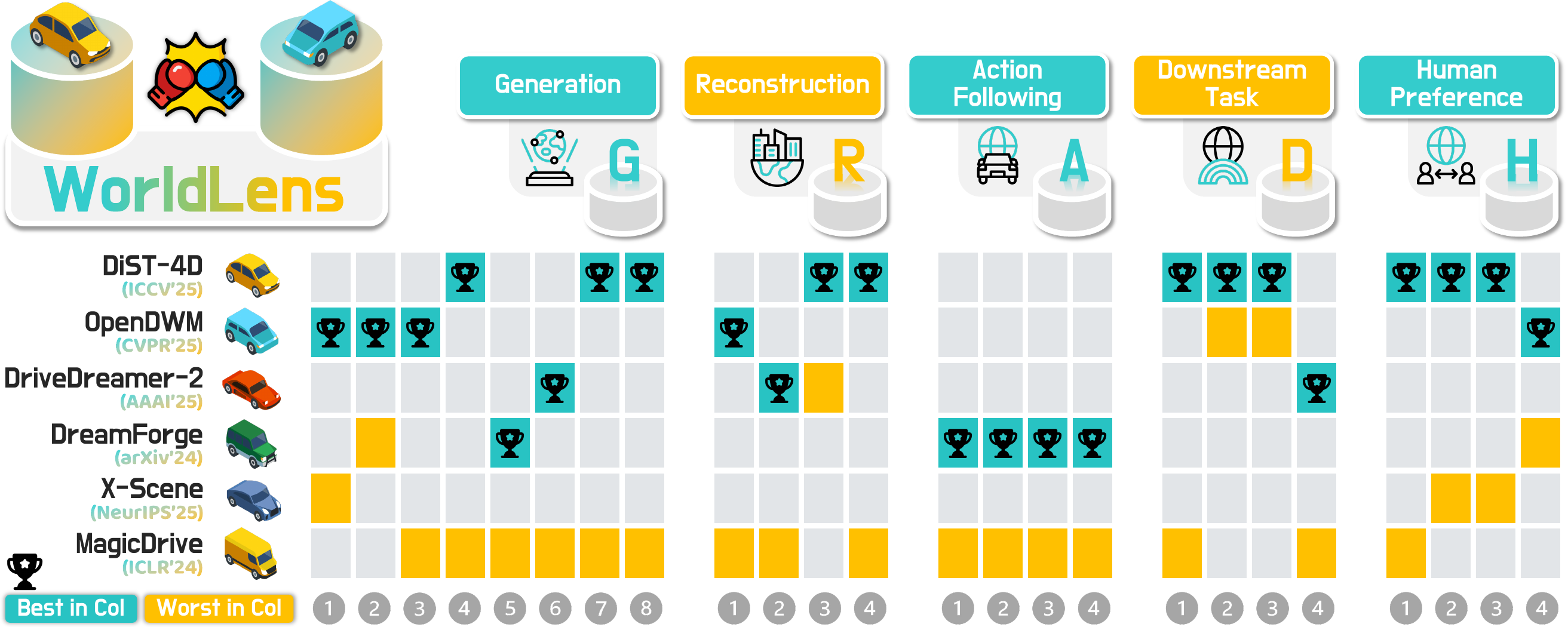}
    \vspace{-0.55cm}
    \caption{\textbf{How do world models perform in the real world?} This work introduces \ours, a unified benchmark for evaluations on $^1$\textbf{\textsl{Generation}}, $^2$\textbf{\textsl{Reconstruction}}, $^3$\textbf{\textsl{Action-Following}}, $^4$\textbf{\textsl{Downstream Task}}, and $^5$\textbf{\textsl{Human Preference}}, across $\mathbf{24}$ \textbf{dimensions}. We observe \emph{no single model dominates across all axes}, highlighting the need for balanced progress toward physically realistic world modeling.}
    \label{fig:teaser}
    \vspace{0.28cm}
    \end{center}
}]

\begin{abstract}
Today's driving world models can generate remarkably realistic dash-cam videos, yet no single model excels universally. Some generate photorealistic textures but violate basic physics; others maintain geometric consistency but fail when subjected to closed-loop planning. This disconnect exposes a critical gap: \emph{the field evaluates how real generated worlds appear, but rarely whether they behave realistically.} We introduce \textbf{\ours}, a unified benchmark that measures world-model fidelity across the full spectrum, from pixel quality and 4D geometry to closed-loop driving and human perceptual alignment, through five complementary aspects and 24 standardized dimensions. Our evaluation of six representative models reveals that no existing approach dominates across all axes: texture-rich models violate geometry, geometry-aware models lack behavioral fidelity, and even the strongest performers achieve only 2--3 out of 10 on human realism ratings. To bridge algorithmic metrics with human perception, we further contribute \textbf{WorldLens-26K}, a 26{,}808-entry human-annotated preference dataset pairing numerical scores with textual rationales, and \textbf{WorldLens-Agent}, a vision-language evaluator distilled from these judgments that enables scalable, explainable auto-assessment. Together, the benchmark, dataset, and agent form a unified ecosystem for assessing generated worlds not merely by visual appeal, but by physical and behavioral fidelity.
\end{abstract}

\vspace{-0.5cm}
\begin{figure*}[t]
    \centering
    \includegraphics[width=\textwidth]{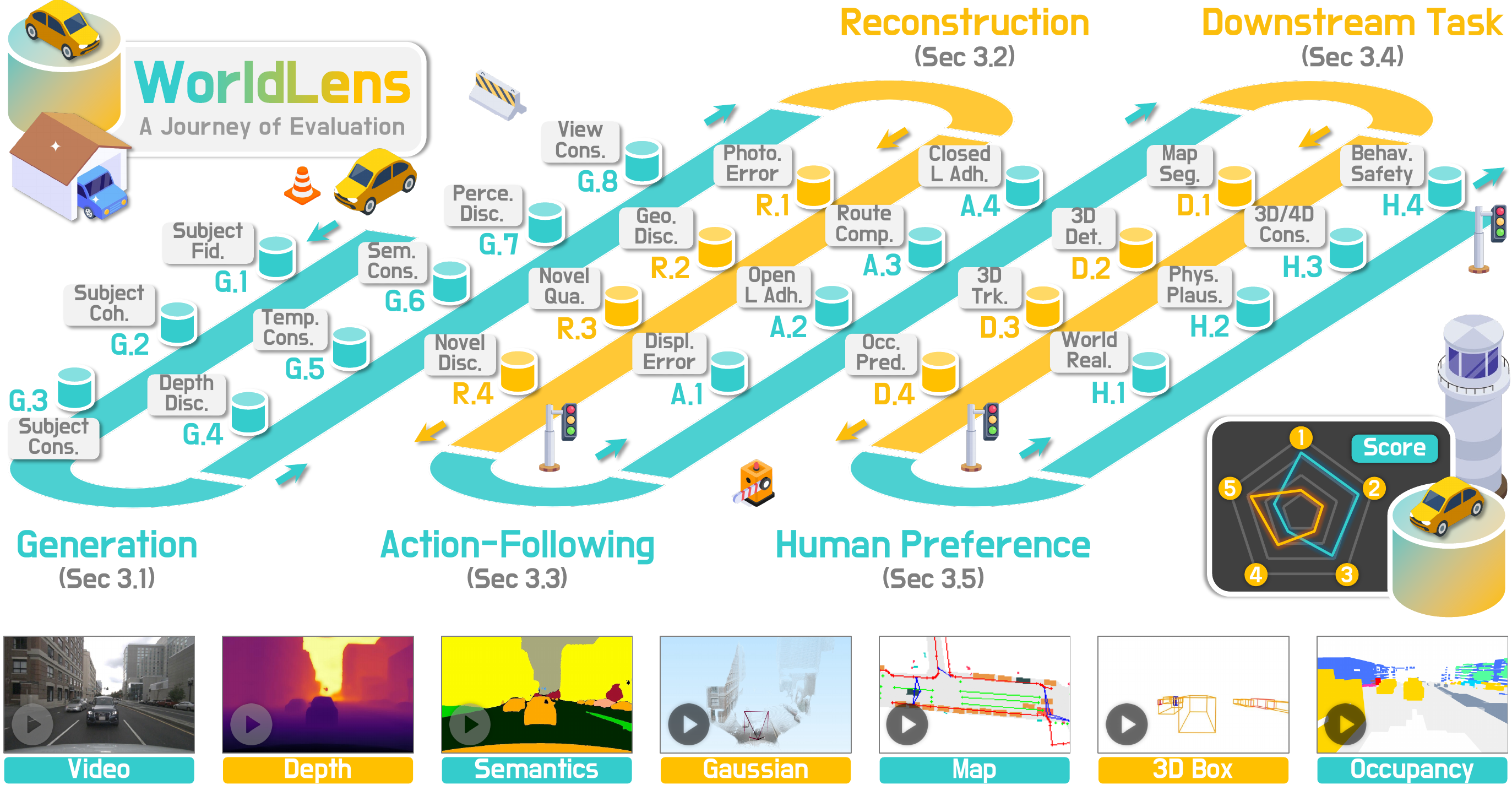}
    \vspace{-0.59cm}
    \caption{\ours~unifies five complementary aspects, namely $^1$\emph{\textbf{Generation}}, $^2$\emph{\textbf{Reconstruction}}, $^3$\emph{\textbf{Action-Following}}, $^4$\emph{\textbf{Downstream Task}}, and $^5$\emph{\textbf{Human Preference}}, that jointly cover visual, structural, functional, and perceptual quality across $24$ interpretable dimensions.}
    \label{fig:bench}
    \vspace{-0.2cm}
\end{figure*}

\section{Introduction}
\label{sec:intro}

Consider a state-of-the-art driving world model that generates visually compelling multi-view videos. The textures are sharp, the lighting is natural, and the motion appears fluid. Yet when a pretrained planner is deployed within this generated world to execute a routine maneuver, it frequently collides or drifts off-road within seconds. This is not an edge case: our experiments show that even the best-performing world models complete fewer than $14\%$ of navigation routes under closed-loop control.

This paradox lies at the heart of the current world-modeling landscape. Recent years have witnessed remarkable progress in generative driving models~\cite{google2024genie2,genie3,russell2025gaia-2}, generating multi-view video sequences that are increasingly difficult to distinguish from real footage. Yet the prevailing evaluation metrics, such as FID, FVD, and LPIPS, were designed for \emph{image quality}, not \emph{world fidelity}~\cite{huang2024vbench,ke2021musiq}. These metrics quantify perceptual similarity but reveal nothing about whether the underlying geometry is coherent, whether the physics are plausible, or whether the generated world can support downstream autonomy tasks~\cite{kong2025survey,onevl}. As such, the field has been optimizing for an incomplete objective: worlds that \emph{appear} real but do not \emph{behave} realistically. The absence of a comprehensive evaluation protocol means that progress on one axis (\eg, texture realism) can mask regression on others (\eg, 3D consistency or action controllability), making it difficult to compare models or identify meaningful bottlenecks.

We introduce \textbf{\ours}, a benchmark designed to address this gap. Rather than scoring generated videos along a single quality axis, we decompose evaluation along \textbf{five complementary axes}: \emph{Generation} (visual realism and temporal stability across eight dimensions), \emph{Reconstruction} (whether videos can be lifted into coherent 4D Gaussian fields), \emph{Action-Following} (whether pretrained planners can operate safely in the generated world), \emph{Downstream Task} (whether synthetic data supports real-world perception models), and \emph{Human Preference} (subjective judgments of realism, physics, and safety from $930$+ hours of annotation). Together, these $24$ dimensions span the full spectrum from pixel fidelity to functional reliability, as illustrated in Figs.~\ref{fig:teaser} and~\ref{fig:bench}. By evaluating each model through every lens simultaneously, WorldLens makes the trade-offs between competing design choices explicit and quantifiable.

Our evaluation reveals a notable finding: \textbf{no existing world model dominates across all axes}. Models that lead in texture quality often violate physics; those with strong geometry fail behaviorally; and even the best performers score only $2$--$3$ out of $10$ on human realism ratings. These results suggest that current architectures still treat appearance, geometry, and dynamics as largely independent objectives, a decomposition that prevents holistic world understanding. To enable such evaluation at scale beyond manual annotation, we further contribute \textbf{WorldLens-26K}, a dataset of $26{,}808$ human-scored video entries with textual rationales, and \textbf{WorldLens-Agent}, a vision-language critic agent distilled from these annotations via LoRA-based SFT on Qwen3-VL-8B~\cite{Qwen2.5-VL}. The agent achieves strong zero-shot alignment with human judgments on unseen models, enabling automated, explainable evaluation at scale.

\section{The \ours~Benchmark}
\label{sec:bench}

A world model that generates visually appealing frames does not necessarily \emph{understand} the world it portrays. To bridge this gap between surface-level appearance and deeper understanding, WorldLens structures evaluation along five complementary axes (Fig.~\ref{fig:bench}), each probing a distinct facet of world fidelity, ranging from low-level pixel quality to high-level behavioral realism.

\noindent\textbf{Generation: How Realistic Does It Appear?}
We decompose visual quality into eight dimensions organized around three questions. \emph{Are individual objects realistic?} Subject Fidelity~(G.1, $\uparrow$) scores cropped objects via class-specific classifiers~\cite{dosovitskiy2020image}; Subject Coherence~(G.2, $\uparrow$) and Consistency~(G.3, $\uparrow$) track identity stability across frames using ReID~\cite{zuo2024cross,he2021transreid} and DINO~\cite{caron2021dino} features. \emph{Is the scene temporally and geometrically stable?} Depth Discrepancy~(G.4, $\downarrow$) measures frame-to-frame depth smoothness \cite{yang2024depth-any-v2}; Temporal (G.5, $\uparrow$) and Semantic (G.6, $\uparrow$) Consistency assess global stability in CLIP~\cite{radford2021learning} and SegFormer~\cite{xie2021segformer} spaces, respectively. \emph{Is the video perceptually convincing as a whole?} Perceptual Discrepancy~(G.7, $\downarrow$) computes FVD on I3D features~\cite{unterthiner2018towards,carreira2017i3d}, and Cross-View Consistency~(G.8, $\uparrow$) evaluates multi-camera alignment via LoFTR~\cite{sun2021loftr}.

\noindent\textbf{Reconstruction: Can a Coherent 3D World Be Recovered?}
This aspect serves as the definitive test for geometric coherence. We lift each generated sequence into a 4D Gaussian field~\cite{kerbl2023-3dgs} and re-render it under both training and novel camera poses. Models generating sharp 2D frames often exhibit severe degradation under this protocol, generating geometric ``floaters" and depth artifacts that reveal how loosely current architectures couple temporal frames. We measure Photometric Error~(R.1, $\downarrow$) via LPIPS/PSNR/SSIM at training poses, Geometric Discrepancy~(R.2, $\downarrow$) via depth comparison using Grounded-SAM 2~\cite{ren2024grounded-sam,ravi2025sam-2}, and both Novel-View Quality~(R.3, $\uparrow$) and Novel-View Discrepancy~(R.4, $\downarrow$) under unseen viewpoints.

\noindent\textbf{Action-Following: Can a Planner Operate Reliably Within It?}
This is arguably the most consequential aspect: if a pretrained planner~\cite{hu2023uniad,jiang2023vad} cannot make safe decisions within the generated world, the model's practical value for autonomous driving remains limited. We measure Displacement Error~(A.1, $\downarrow$) between predicted and real trajectories, Open-Loop Adherence~(A.2, $\uparrow$) via the PDMS score~\cite{dauner2024navsim}, Route Completion~(A.3, $\uparrow$) in closed-loop simulation, and Closed-Loop Adherence~(A.4, $\uparrow$) via the Arena Driving Score~\cite{yang2024drivearena}. As we show in Sec.~\ref{sec:experiments}, the disparity between open- and closed-loop results is substantial and highly informative.

\noindent\textbf{Downstream Task: Is the Synthetic Data Practically Useful?}
We assess whether generated videos can support 3D perception pipelines trained on real data. Specifically, we evaluate Map Segmentation~(D.1, $\uparrow$) via BEVFusion~\cite{liu2023bevfusion}, 3D Object Detection~(D.2, $\uparrow$) via NDS~\cite{caesar2020nuscenes}, 3D Tracking~(D.3, $\uparrow$) via AMOTA~\cite{ding2024ada}, and Occupancy Prediction~(D.4, $\uparrow$) via SparseOcc RayIoU~\cite{tang2024sparseocc}. Notably, even models with strong perceptual quality can degrade detection accuracy by $30$--$50\%$.

\begin{figure}[t]
    \centering
    \includegraphics[width=\linewidth]{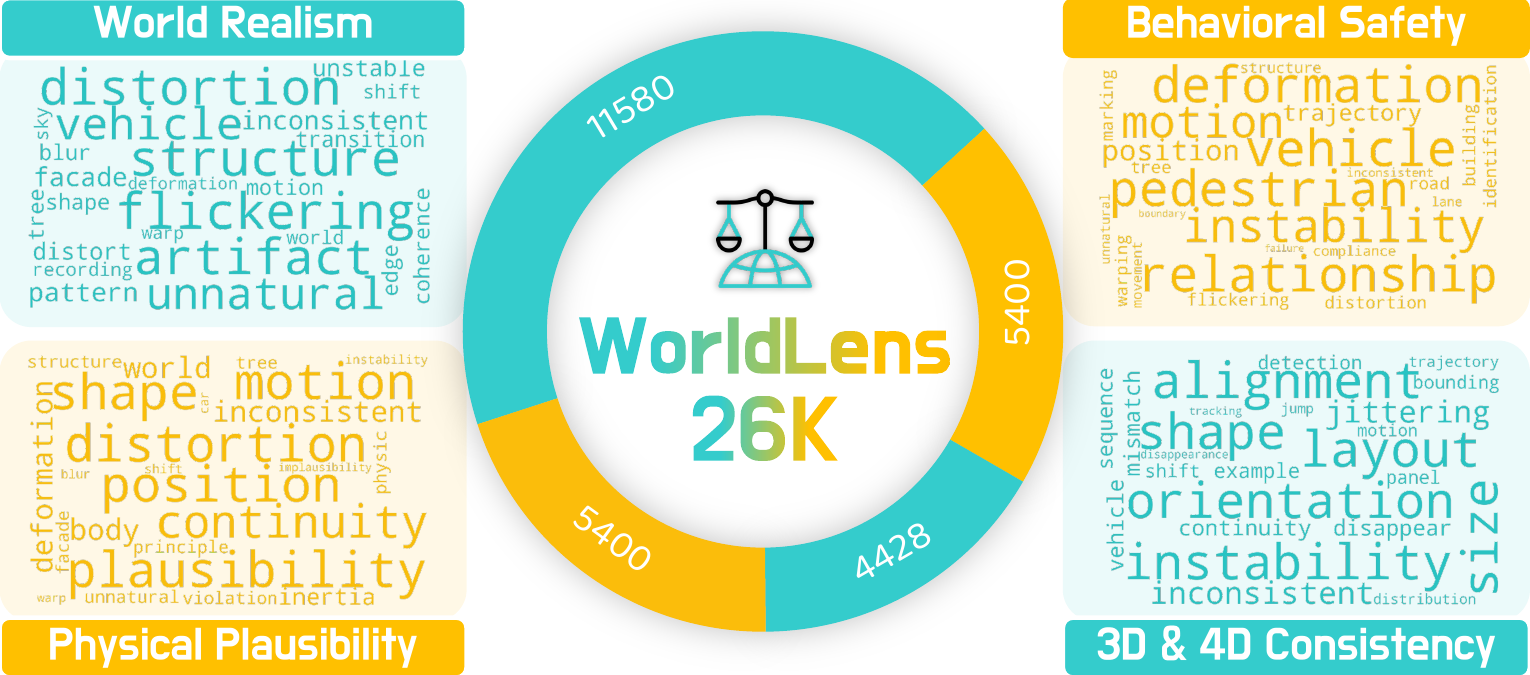}
    \vspace{-0.6cm}
    \caption{
    Statistics and word clouds of \textbf{WorldLens-26K}. Frequent keywords align with target criteria, confirming that annotators maintain consistent, dimension-specific reasoning.
    }
    \label{fig:dataset}
    \vspace{-0.3cm}
\end{figure}

\begin{table*}[t]
    \centering
    \caption{Summary of benchmarking results of state-of-the-art driving world models for \textbf{Generation} and \textbf{Reconstruction} in WorldLens. }
    \vspace{-0.2cm}
    \label{tab:bench_gen_recon}
    \resizebox{\linewidth}{!}{
    \begin{tabular}{r|r|cccccccc|cccc}
        \toprule
        & & \multicolumn{8}{c|}{\cellcolor{w_blue!24}\textbf{~Aspect: Generation}} & \multicolumn{4}{c}{\cellcolor{w_yellow!24}\textbf{~Aspect: Reconstruction}}
        \\
        \multirow{2.2}{*}{\textbf{Model}} & \multirow{2.2}{*}{\textbf{Venue}} & Subject & Subject & Subject & Depth & Temp. & Sem. & Percept. & View & Photo. & Geo. & Novel & Novel
        \\
        & & ~Fid.$\uparrow$ & ~Cohe.$\uparrow$ & ~Cons.$\uparrow$ & ~Disc.$\downarrow$ & ~Cons.$\uparrow$ & ~Cons.$\uparrow$ & ~Disc.$\downarrow$ & ~Cons.$\uparrow$ & ~Error$\downarrow$ & ~Disc.$\downarrow$ & ~Qual.$\uparrow$ & ~Disc.$\downarrow$
        \\
        & & {{\raisebox{0.9pt}{\colorbox{w_blue}{\scriptsize{\textbf{\textbf{\textsc{\textcolor{white}{~G.1~}}}}}}}}} & {{\raisebox{0.9pt}{\colorbox{w_blue}{\scriptsize{\textbf{\textbf{\textsc{\textcolor{white}{~G.2~}}}}}}}}} & {{\raisebox{0.9pt}{\colorbox{w_blue}{\scriptsize{\textbf{\textbf{\textsc{\textcolor{white}{~G.3~}}}}}}}}} & {{\raisebox{0.9pt}{\colorbox{w_blue}{\scriptsize{\textbf{\textbf{\textsc{\textcolor{white}{~G.4~}}}}}}}}} & {{\raisebox{0.9pt}{\colorbox{w_blue}{\scriptsize{\textbf{\textbf{\textsc{\textcolor{white}{~G.5~}}}}}}}}} & {{\raisebox{0.9pt}{\colorbox{w_blue}{\scriptsize{\textbf{\textbf{\textsc{\textcolor{white}{~G.6~}}}}}}}}} & {{\raisebox{0.9pt}{\colorbox{w_blue}{\scriptsize{\textbf{\textbf{\textsc{\textcolor{white}{~G.7~}}}}}}}}} & {{\raisebox{0.9pt}{\colorbox{w_blue}{\scriptsize{\textbf{\textbf{\textsc{\textcolor{white}{~G.8~}}}}}}}}} & {{\raisebox{0.9pt}{\colorbox{w_yellow}{\scriptsize{\textbf{\textbf{\textsc{\textcolor{white}{~R.1~}}}}}}}}} & {{\raisebox{0.9pt}{\colorbox{w_yellow}{\scriptsize{\textbf{\textbf{\textsc{\textcolor{white}{~R.2~}}}}}}}}} & {{\raisebox{0.9pt}{\colorbox{w_yellow}{\scriptsize{\textbf{\textbf{\textsc{\textcolor{white}{~R.3~}}}}}}}}} & {{\raisebox{0.9pt}{\colorbox{w_yellow}{\scriptsize{\textbf{\textbf{\textsc{\textcolor{white}{~R.4~}}}}}}}}}
        \\
        \midrule
        MagicDrive \cite{gao2023magicdrive} & {\small ICLR'23} & $28.49$ & $75.95$ & $65.22\%$ & $24.19$ & $74.44\%$ & $80.63\%$ & $222.00$ & $185.77$ & $0.140$ & $0.115$ & $39.82\%$ & $427.30$
        \\
        DreamForge \cite{mei2024dreamforge} & {\small arXiv'24} & \underline{$31.99$} & $75.12$ & \underline{$76.40\%$} & $19.27$ & \cellcolor{w_blue!20}$\mathbf{79.82\%}$ & \underline{$84.99\%$} & $189.76$ & $194.99$ & $0.097$ & $0.105$ & $\underline{41.23\%}$ & $347.70$
        \\
        DriveDreamer-2 \cite{zhao2024drivedreamer-2} & {\small AAAI'25} & $27.38$ & $78.97$ & $74.49\%$ & \underline{$17.73$} & $79.51\%$ & \cellcolor{w_blue!20}$\mathbf{85.91\%}$ & $127.07$ & \underline{$302.83$} & $0.093$ & \cellcolor{w_yellow!20}$\mathbf{0.073}$ & $36.10\%$ & $\underline{259.91}$
        \\
        OpenDWM \cite{opendwm} & {\small CVPR'25} & \cellcolor{w_blue!20}$\mathbf{36.30}$ & \cellcolor{w_blue!20}$\mathbf{83.13}$ & \cellcolor{w_blue!20}$\mathbf{78.33\%}$ & $18.17$ & \underline{$79.63\%$} & $84.08\%$ & \underline{$90.42$} & $211.18$ & \cellcolor{w_yellow!20}$\mathbf{0.065}$ & $0.088$ & $39.54\%$ & $287.73$
        \\
        DiST-4D \cite{guo2025dist-4d} & {\small ICCV'25} & $30.32$ & \underline{$79.36$} & $74.69\%$ & \cellcolor{w_blue!20}$\mathbf{17.71}$ & $77.76\%$ & $84.32\%$ & \cellcolor{w_blue!20}$\mathbf{58.08}$ & \cellcolor{w_blue!20}$\mathbf{389.78}$ & $\underline{0.066}$ & $\underline{0.080}$ & \cellcolor{w_yellow!20}$\mathbf{43.09\%}$ & \cellcolor{w_yellow!20}$\mathbf{192.39}$
        \\
        $\mathcal{X}$-Scene \cite{yang2025x-scene} & {\small NeurIPS'25} & $27.17$ & $77.22$ & $74.37\%$ & $20.50$ & $79.41\%$ & $83.80\%$ & $179.74$ & $201.00$ & $0.098$ & $0.096$ & $38.04\%$ & $365.71$
        \\
        \midrule
        \rowcolor{gray!7}\textcolor{gray}{Empirical Max} & \textcolor{gray}{-} & \textcolor{gray}{$60.22$} & \textcolor{gray}{$83.25$} & \textcolor{gray}{$93.66\%$} & \textcolor{gray}{$14.27$} & \textcolor{gray}{$93.24\%$} & \textcolor{gray}{$86.39\%$} & \textcolor{gray}{-} & \textcolor{gray}{$570.75$} & \textcolor{gray}{0.056} & \textcolor{gray}{-} & \textcolor{gray}{$45.69\%$} & \textcolor{gray}{-}
        \\
        \bottomrule
    \end{tabular}}
    \vspace{-0.1cm}
\end{table*}

\begin{figure*}[t]
    \centering
    \includegraphics[width=\textwidth]{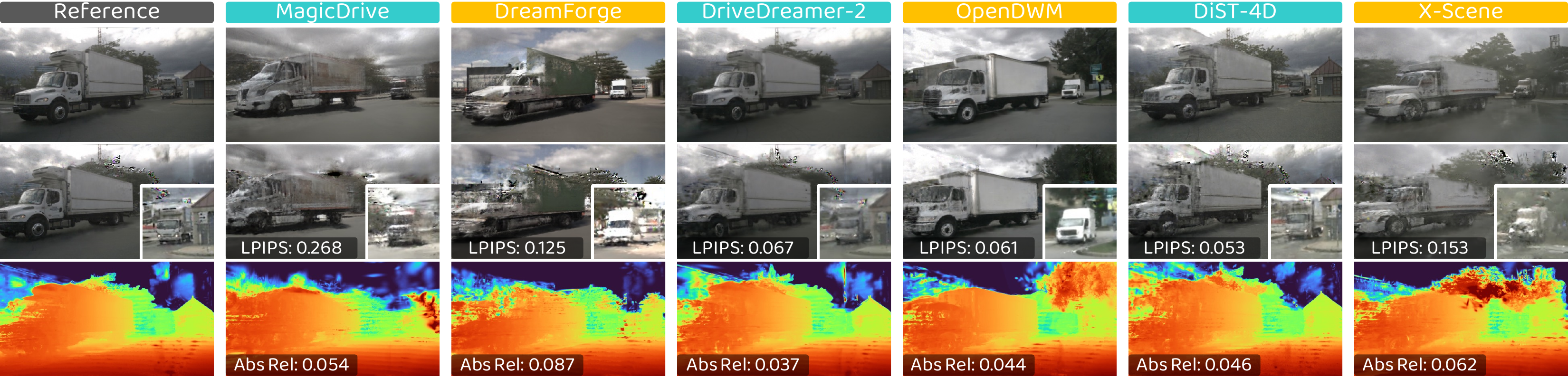}
    \vspace{-0.6cm}
    \caption{
        \textbf{4D reconstruction} from generated videos. Rows: $^1$generated frame, $^2$novel-view rendering at a lateral offset, $^3$depth map.
    }
    \label{fig:vis_recon}
    \vspace{-0.3cm}
\end{figure*}

\begin{figure*}[t]
    \centering
    \includegraphics[width=\textwidth]{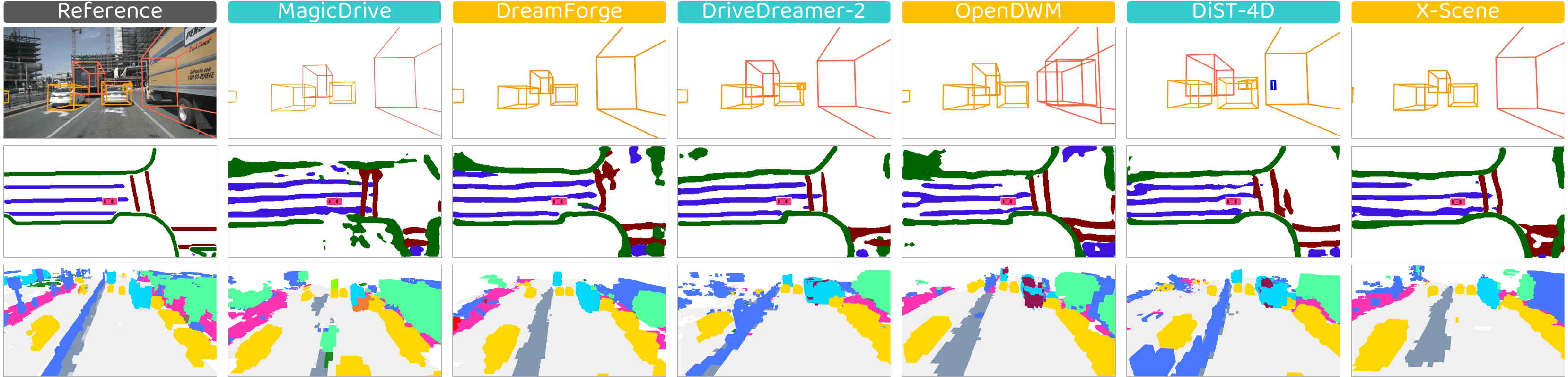}
    \vspace{-0.65cm}
    \caption{
        \textbf{Downstream task} qualitative results. Rows: $^1$3D detection, $^2$BEV map segmentation, and $^3$semantic occupancy prediction.
    }
    \label{fig:vis_downstream}
    \vspace{-0.2cm}
\end{figure*}

\noindent\textbf{Human Preference: How Do Human Observers Judge Quality?}
Quantitative metrics can miss perceptual artifacts that human observers identify readily. Ten annotators in two independent groups rated generated videos on a $1$--$10$ scale across four perceptual dimensions: World Realism~(H.1), Physical Plausibility~(H.2), Behavioral Safety~(H.4), and 3D \& 4D Consistency~(H.3). Each annotator reviewed four time-synchronized modalities, including the generated video alongside its semantic segmentation mask, estimated depth map, and 3D bounding box overlay, facilitating holistic judgment across visual, structural, and behavioral aspects. Each annotation required ${\sim}$128 seconds on average, totaling over $\mathbf{930}$ hours. Divergent ratings between groups were re-evaluated to ensure inter-annotator consistency.

\noindent\textbf{From Human Judgment to Scalable Evaluation.}

The annotation effort leads to \textbf{WorldLens-26K}, a dataset of 26{,}808 scoring records, each coupling a numerical rating with a free-text rationale that articulates \emph{why} a particular score was assigned. As in Fig.~\ref{fig:dataset}, word-cloud of these rationales reveals strong topical alignment with their respective dimensions (\eg, ``shape" and ``reflection" for realism, ``motion" and ``safety" for behavioral assessment), confirming that annotators systematically attend to the intended perceptual criteria. Building on this paired quantitative--qualitative supervision, we train \textbf{WorldLens-Agent}, a vision-language evaluator distilled from human preferences through LoRA-based SFT on Qwen3-VL~\cite{Qwen2.5-VL}. The agent simultaneously predicts dimension-specific scores and produces free-text explanations that mirror human reasoning patterns, demonstrating strong zero-shot generalization to previously unseen models (Fig.~\ref{fig:agent}) and offering a practical pathway toward scalable, interpretable auto-evaluation without ongoing manual labeling.
\section{Key Findings}
\label{sec:experiments}

We evaluate six representative models: MagicDrive~\cite{gao2023magicdrive}, DreamForge~\cite{mei2024dreamforge}, DriveDreamer-2~\cite{zhao2024drivedreamer-2}, OpenDWM~\cite{opendwm}, DiST-4D~\cite{guo2025dist-4d}, and $\mathcal{X}$-Scene~\cite{yang2025x-scene}, across all five aspects. Rather than enumerating each metric individually, we organize the results around the most salient insights. Comprehensive details are available in the main paper~\cite{liang2026worldlens}.

\noindent\emph{\textbf{No model is an all-rounder.}}
The radar chart in Fig.~\ref{fig:teaser} summarizes the landscape concisely: every model exhibits notable weaknesses. OpenDWM~\cite{opendwm} leads in subject fidelity and coherence owing to multi-dataset training, yet DiST-4D~\cite{guo2025dist-4d} dominates in perceptual discrepancy, cross-view consistency, and all reconstruction metrics (\cref{tab:bench_gen_recon}). DriveDreamer-2~\cite{zhao2024drivedreamer-2} achieves the strongest semantic consistency and geometric discrepancy, but falls short in subject-level realism. All models remain substantially below the Empirical Max, indicating that the frontier of driving world generation is far from saturated.

\noindent\emph{\textbf{Perceptual quality $\neq$ functional utility.}}
Perhaps the most significant finding is the disconnect between perceptual quality and downstream utility. OpenDWM achieves the strongest subject fidelity ($36.30$) and the lowest photometric reconstruction error ($0.065$ LPIPS), yet it scores only $21.96\%$ on 3D object detection and $6.90\%$ on tracking, roughly $30$--$50\%$ below DiST-4D. This reveals that large-scale multi-domain training can enhance visual diversity while simultaneously hindering alignment with task-specific data distributions. In contrast, DiST-4D's geometry-aware RGB-D generation trades some visual fidelity for improved downstream performance ($35.55\%$ map segmentation, $33.22\%$ detection, $15.30\%$ tracking).

\noindent\emph{\textbf{2D fidelity does not guarantee 4D coherence.}}
Reconstruction serves as the definitive test, and most models exhibit significant degradation. When generated videos are lifted into 4D Gaussian fields and re-rendered from novel viewpoints, MagicDrive~\cite{gao2023magicdrive} generates dense floaters and over $2\times$ higher photometric error compared to OpenDWM (Fig.~\ref{fig:vis_recon}). DreamForge~\cite{mei2024dreamforge} exhibits similar artifacts. Only DiST-4D maintains clean, stable geometry under lateral camera shifts, with its decoupled spatiotemporal diffusion and explicit depth supervision cutting photometric and geometric error by ${\sim}55\%$. This reveals a fundamental limitation: most diffusion-based architectures couple temporal frames too loosely to achieve the spatial consistency required for faithful 3D reconstruction.

\noindent\emph{\textbf{Open-loop performance masks closed-loop failure.}}
All evaluated models achieve reasonable open-loop PDMS scores ($71\%$--$79\%$), indicating that planners can extract meaningful cues from generated frames in isolation. However, under closed-loop control, where planning decisions feed back into the simulation, route completion rates drop sharply to $6$--$14\%$. The best-performing model, RLGF~\cite{yan2025rlgf}, completes only $13.51\%$ of routes. Frequent collisions and off-road departures confirm that photorealistic appearance alone cannot sustain safe navigation; the generated worlds lack the causal and physical consistency required for sustained autonomous control.

\begin{figure}[t]
    \centering
    \begin{subfigure}[t]{0.48\linewidth}
        \centering
        \includegraphics[width=\linewidth]{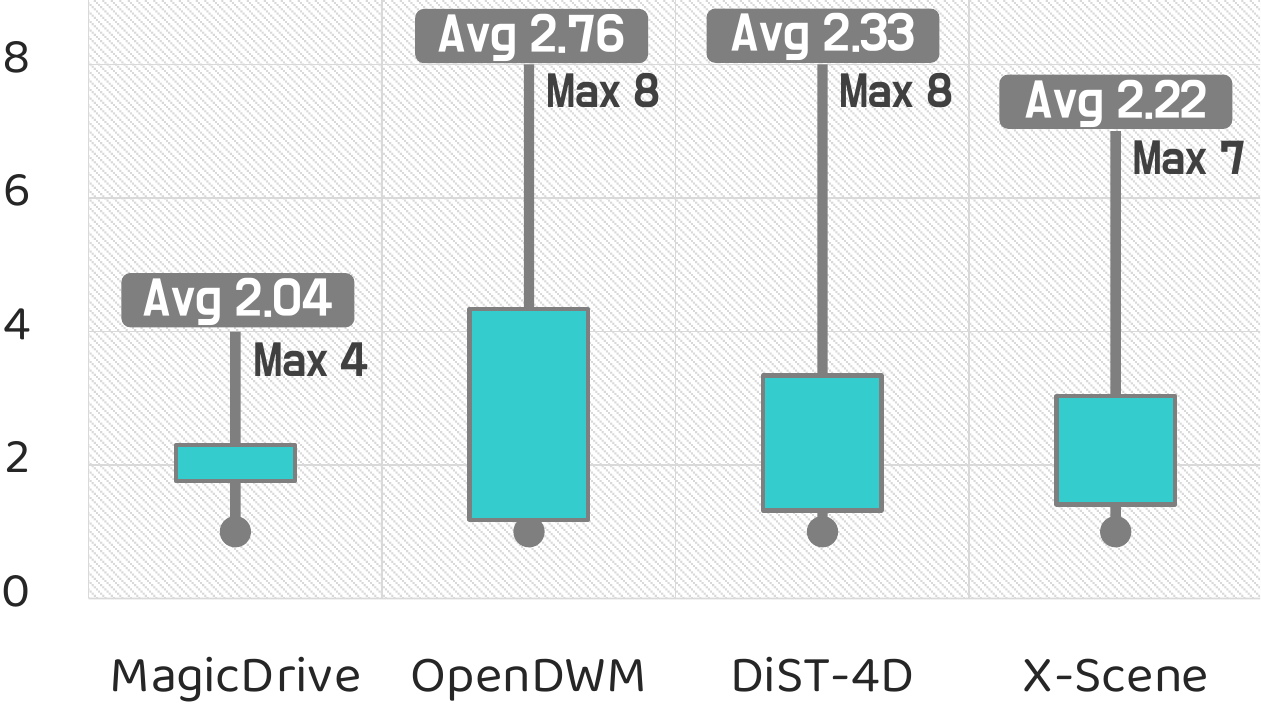}
        \caption{World Realism}
    \end{subfigure}
    \hfill
    \begin{subfigure}[t]{0.48\linewidth}
        \centering
        \includegraphics[width=\linewidth]{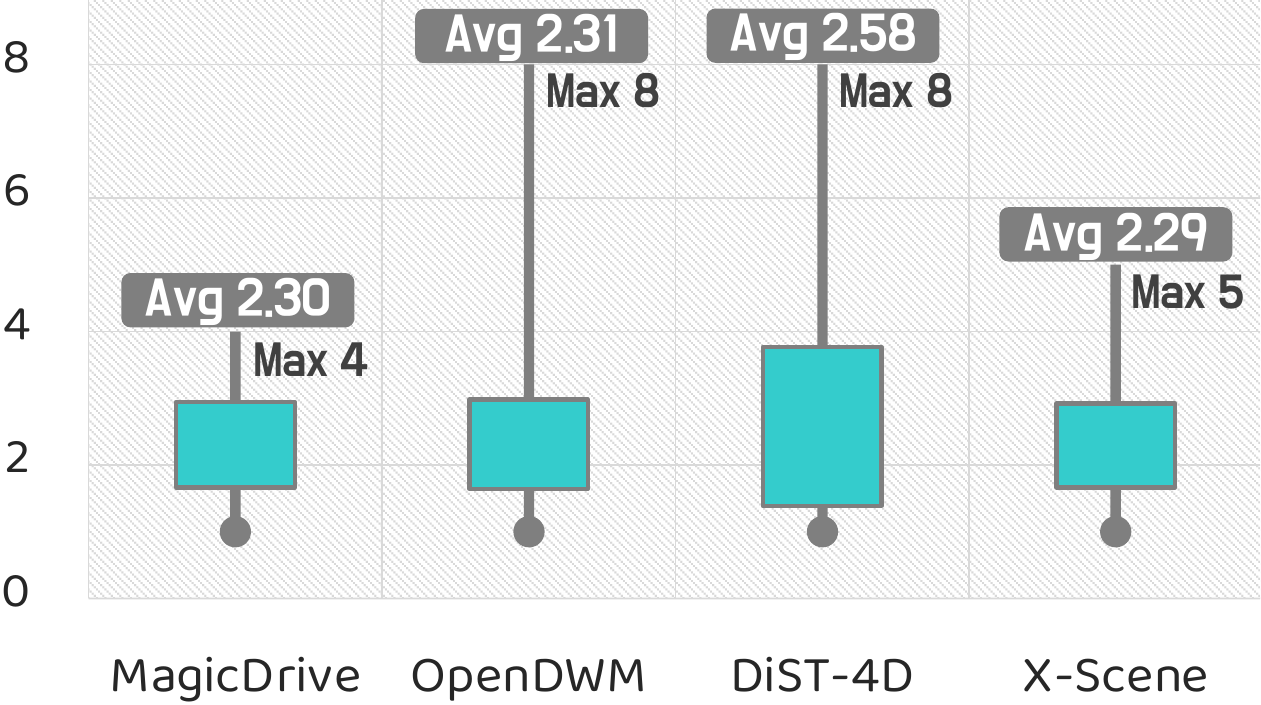}
        \caption{Physical Plausibility}
    \end{subfigure}
    \vspace{3pt}
    \begin{subfigure}[t]{0.48\linewidth}
        \centering
        \includegraphics[width=\linewidth]{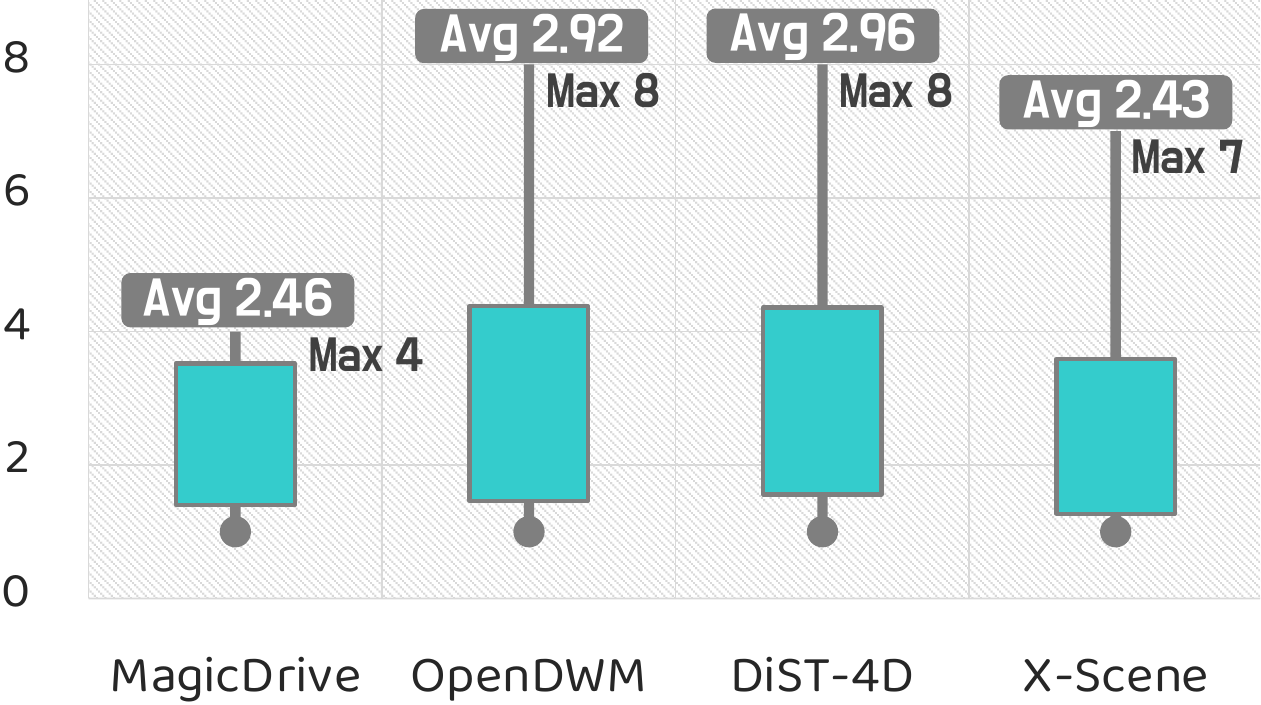}
        \caption{3D \& 4D Consistency}
    \end{subfigure}
    \hfill
    \begin{subfigure}[t]{0.48\linewidth}
        \centering
        \includegraphics[width=\linewidth]{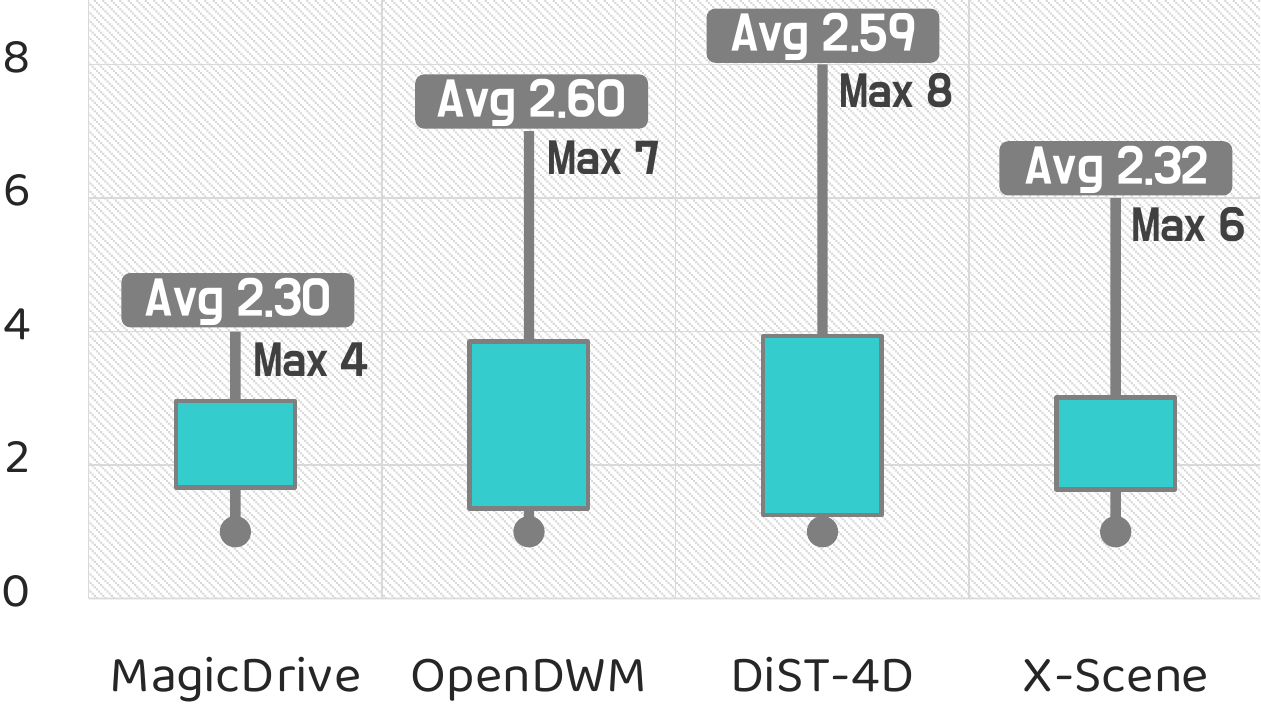}
        \caption{Behavioral Safety}
    \end{subfigure}
    \vspace{-0.3cm}
    \caption{\textbf{Human Preference} alignments. Max, median, and average scores for each model across four perceptual dimensions. All scores remain modest ($2$--$3$ out of $10$), with geometric consistency strongly correlated with perceived realism.}
    \label{fig:human_preference}
    \vspace{-0.2cm}
\end{figure}

\noindent\emph{\textbf{Human perception correlates with geometric consistency.}}
Human preference scores remain notably low: averages of $2$--$3$ out of $10$ across all dimensions (Fig.~\ref{fig:human_preference}), underscoring that existing world models still fall well short of human-perceived realism. However, a meaningful pattern emerges: models with stronger geometric consistency (\eg, DiST-4D) receive higher world realism and physical plausibility ratings, while perceptually appealing but geometrically unstable models (\eg, MagicDrive) score lowest. The strong correlation between consistency and perceived realism indicates that human observers are inherently sensitive to structural coherence, which underscores the importance of geometry-aware evaluation and training.

\begin{figure}[t]
    \centering
    \includegraphics[width=0.99\linewidth]{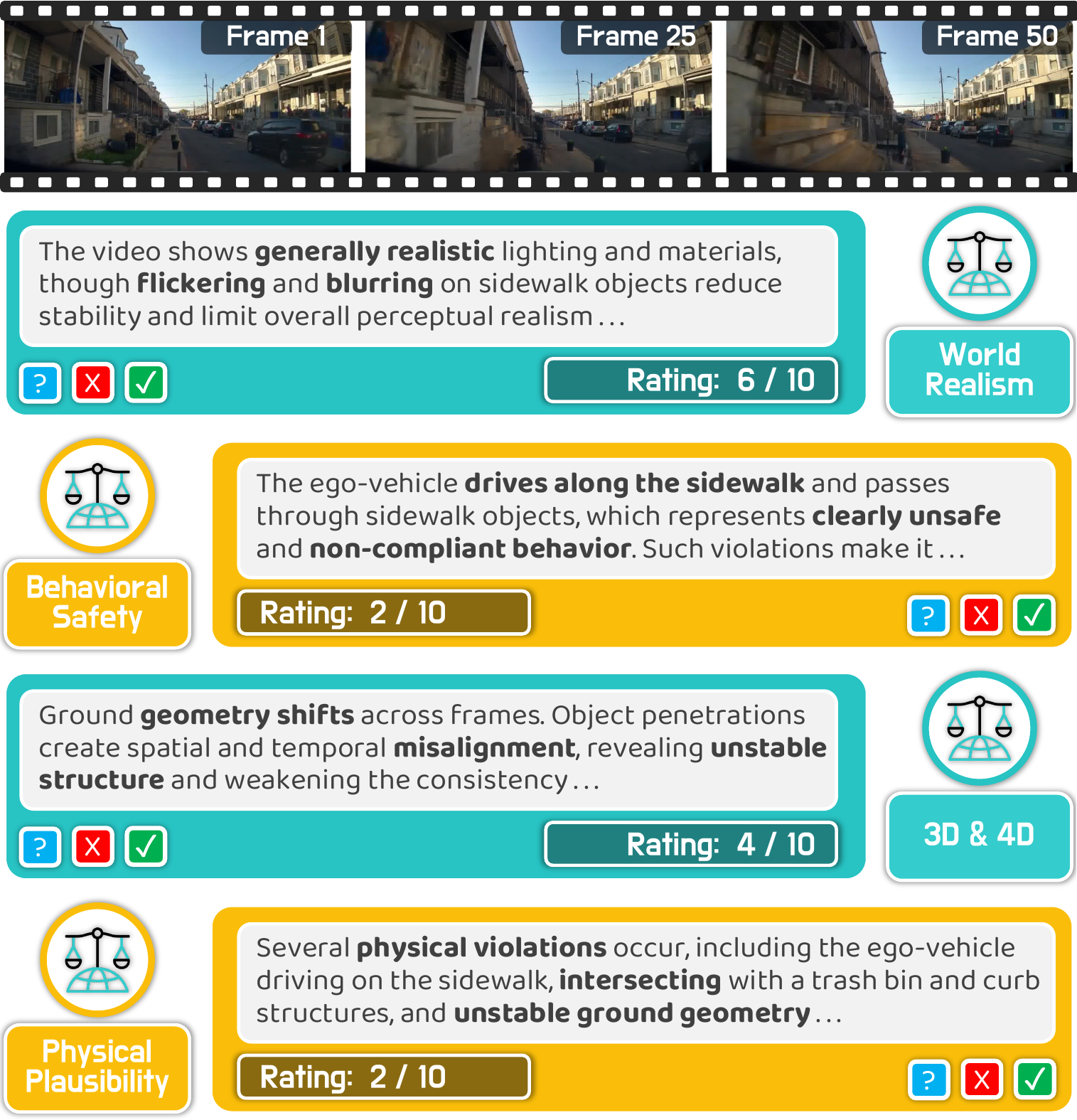}
    \vspace{-0.2cm}
    \caption{
    Zero-shot evaluations by \textbf{WorldLens-Agent} on unseen videos (from Gen3C~\cite{ren2025gen3c}), exhibiting strong alignment with human scores and reasoning.}
    \label{fig:agent}
    \vspace{-0.25cm}
\end{figure}

\noindent\emph{\textbf{WorldLens-Agent generalizes to unseen models.}}
To validate the automated evaluator, we perform zero-shot assessment on Gen3C~\cite{ren2025gen3c} videos, which were never seen during training. As in Fig.~\ref{fig:agent}, the agent's ratings closely track human annotations, and its generated rationales faithfully reflect annotator reasoning, confirming reliable score-level agreement and coherent interpretive alignment. This validates the feasibility of encoding human perceptual standards into a scalable, automated evaluation pipeline.

\noindent\textbf{Design Principles for Future World Models.}

These findings converge on four actionable principles:

\noindent\emph{(1)} Treat geometry as a first-class training objective, as explicit depth supervision consistently improves both reconstruction and downstream perception. 

\noindent\emph{(2)} Optimize jointly for appearance and structure, since current pipelines that decouple texture from geometry incur significant degradation in 4D consistency. 

\noindent\emph{(3)} Evaluate under closed-loop conditions, not solely open-loop, because the open/closed-loop disparity is the most reliable indicator of world-model maturity. 

\noindent\emph{(4)} Benchmark comprehensively, as single-axis metrics systematically overlook the trade-offs that characterize the current landscape.

\section{Conclusion}
\label{sec:conclusion}

WorldLens reveals a critical reality: the driving world models widely recognized for visual realism remain far from functional deployment. No model dominates across all evaluation axes, human ratings average only $2$--$3$ out of $10$, and closed-loop control degrades almost universally. These deficiencies are not incremental; they point to fundamental gaps in how generative worlds are \emph{constructed} and \emph{evaluated}. By unifying $24$ evaluation dimensions, we provide the community with a standardized, scalable, and human-aligned protocol for measuring progress toward worlds that are physically coherent, functionally reliable, and suitable for autonomous navigation. We anticipate that this benchmark will accelerate the transition from appearance-driven generation to behavior-grounded world modeling.

{
    \small
    \bibliographystyle{ieeenat_fullname}
    \bibliography{main}
}

\end{document}